%% file: tacl.tex
\PassOptionsToPackage{table}{xcolor}

\documentclass[11pt,a4paper]{article}
\usepackage{times,latexsym}
\usepackage{url}
\usepackage[T1]{fontenc}
\usepackage{amsmath}
\usepackage{graphicx}
\usepackage{enumitem}
\usepackage{geometry}
\usepackage{array}
\usepackage{xcolor}
\usepackage{colortbl}
\usepackage{ragged2e}
\usepackage{svg} 

\usepackage{caption}
\usepackage{xcolor}

\usepackage{titlesec}

\usepackage[table]{xcolor} 
\usepackage{tabularx}       
\usepackage{geometry}       

\titlespacing*{\section}{0pt}{0.5ex}{0.5ex} 
\titlespacing*{\subsection}{0pt}{0.5ex}{0.5ex} 
\titlespacing*{\subsubsection}{0pt}{0.5ex}{0.5ex} 
\usepackage[acceptedWithA]{tacl2021v1}
\usepackage{xspace,mfirstuc,tabulary}

\newif\iftaclinstructions
\taclinstructionsfalse 
\iftaclinstructions
\renewcommand{\confidential}{}
\renewcommand{\anonsubtext}{(No author info supplied here, for consistency with
TACL-submission anonymization requirements)}
\newcommand{\instr}
\fi

\iftaclpubformat 

\else

\fi

\title{A Practical Guide for Evaluating LLMs and LLM-Reliant Systems}

\author{
  Ethan M. Rudd
  \\
  Google
  \\
  \texttt{ethanrudd@google.com}
  \And
 Christopher Andrews 
  \\
  Google
  \\
  \texttt{andrewsch@google.com}
  \And
    Philip Tully 
    \\
    Google
  \\
  \texttt{phtully@google.com}
}

\definecolor{headercolor}{gray}{0.85}
\definecolor{promptcolor}{rgb}{0.87, 0.92, 0.98}   
\definecolor{goldencolor}{rgb}{0.9, 1.0, 0.9}      
\definecolor{errorcolor}{rgb}{1.0, 0.85, 0.85}     
\definecolor{paraphrasecolor}{rgb}{0.95, 1.0, 0.85}
\definecolor{incompletecolor}{rgb}{1.0, 0.95, 0.8} 
\definecolor{irrelevantcolor}{rgb}{0.95, 0.9, 1.0} 

\vspace{-2em}
\begin{document}
\maketitle
\begin{abstract}
Recent advances in generative AI have led to remarkable interest in using systems that rely on large language models (LLMs) for practical applications. However, meaningful evaluation of these systems in real-world scenarios comes with a distinct set of challenges, which are not well-addressed by synthetic benchmarks and de-facto metrics that are often seen in the literature. We present a practical evaluation framework which outlines how to proactively curate representative datasets, select meaningful evaluation metrics, and employ meaningful evaluation methodologies that integrate well with practical development and deployment of LLM-reliant systems that must adhere to real-world requirements and meet user-facing needs.
\end{abstract}

\input{content/introduction}

\input{content/datasets}

\input{content/metrics}
\input{content/eval}
\input{content/conclusion}
\input{content/acknowledgements}

\bibliography{tacl}
\bibliographystyle{acl_natbib}








  

\end{document}

%% file: content/introduction.tex
\section{Introduction}

Generative AI systems that rely on large language models (LLMs) have seen remarkable adoption, yet their evaluation remains a significant challenge. Users can supply infinitely many prompts, and the systems can in turn output infinitely many responses, often unpredictably. This is made even more challenging for multi-turn agentic workflows where errors can compound with each additional agent execution. Complexity is further exacerbated by sensitivity to minor prompt changes, hallucinations, undesired refusals to respond, and inherent non-determinism stemming from stochastic computations, chained model calls, and grounding data/API inconsistencies. Deploying these systems reliably requires robust methods for measuring response quality against real-world requirements.

\begin{figure*}[!t]
  \centering
  \includegraphics[width=\textwidth]{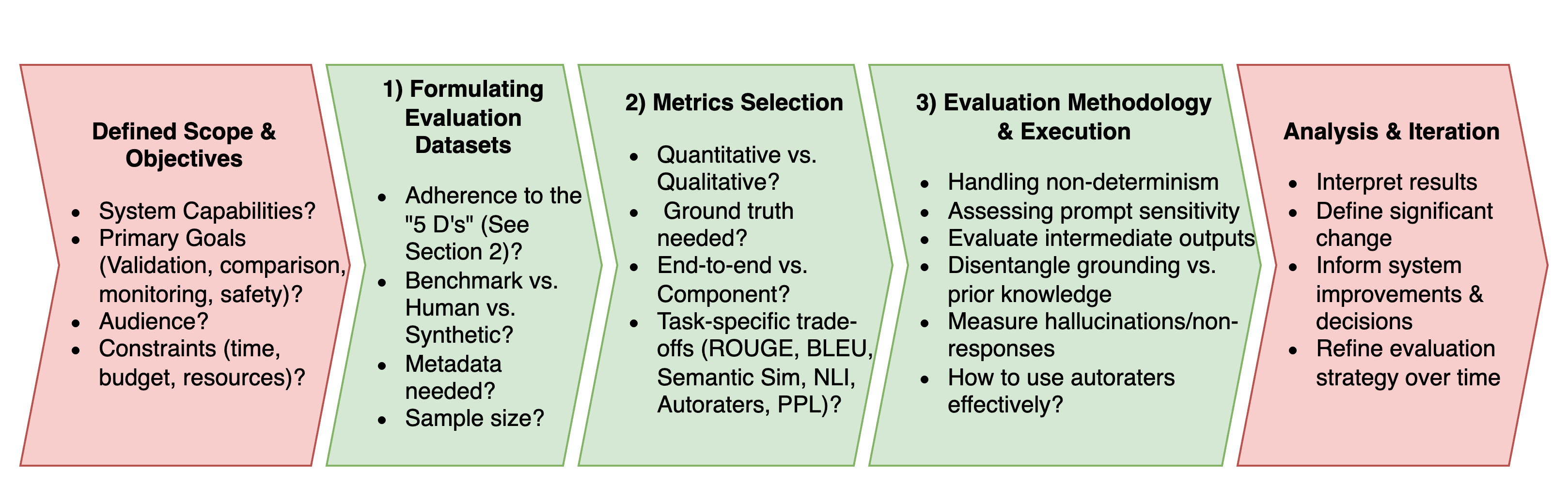} 
  \caption{\textit{Proposed framework for developing an LLM evaluation strategy.} This flowchart guides decisions across three key stages, aligning with the paper's focus on Datasets, Metrics, and Methodology. These stages are shown in green. Additional inputs and actions to inform and guide system development based on these stages are shown in red. Key factors to consider and questions to ask at each stage of the evaluation design are listed as bullet points.}
  \label{fig:teaser}
\end{figure*}

While the importance of evaluation is clear – enabling iterative progress, building user trust, ensuring consistency, and improving efficiency  – practitioners often select from a plethora of "de-facto" evaluation techniques without a consistent implementation strategy to address real-world evaluation goals, system requirements, or end-user experience. 
This paper addresses this gap by proposing a structured, actionable framework for designing and implementing evaluation of LLM-reliant AI systems (cf. Figure \ref{fig:teaser}). The evaluation design process is organized around three fundamental pillars:

\begin{enumerate}
\item Datasets: Curating representative and high-quality data tailored to the evaluation goals.
\item Metrics: Selecting appropriate quantitative and qualitative measures to assess performance against specified objectives.
\item Methodology: Designing the overall evaluation approach, including specific strategies to handle challenges like non-determinism, prompt sensitivity, and hallucination measurement.
\end{enumerate}

The aim of this framework is to assist in proactively designing evaluation suites that integrate with the development and deployment lifecycles of generative AI systems. By defining scope and objectives up-front, these evaluation suites can be tailored around operational goals and requirements, and through continuous evaluation and iterative refinement, the evaluation suites can mature with their corresponding generative AI systems.

In the following sections, we expand on the high-level flowchart presented in Figure \ref{fig:teaser}, outlining key stages and decision points in developing an evaluation plan. This structured framework serves to equip researchers and practitioners with the tools needed to develop robust, reliable, and informative evaluation strategies.

%% file: content/datasets.tex
\section{Evaluation Dataset Formulation}

Datasets are composed of prompts and optionally associated ground truth responses, depending on the chosen evaluation metrics and methodology. To create a robust and reliable evaluation dataset, five core principles, the {\bf 5 D's}, should be followed. In accordance with translating high-level goals that go into our framework, the dataset must have a {\bf 1.  Defined Scope} that aligns with specific tasks the model is meant to perform. The dataset should further be {\bf 2.  Demonstrative of Production Usage}, mimicking the inputs and scenarios expected from actual users. The dataset should be {\bf 3. Diverse}, capturing the variety of the problem space to avoid a narrow or biased evaluation. To ensure the integrity of the results, the dataset must be {\bf 4. Decontaminated}, meaning it is distinct from any data used during model training to prevent misleadingly high performance metrics. Finally, the dataset should be {\bf 5. Dynamic}, treated as a living body of work that evolves as the real-world application changes.

\subsection{Amassing Evaluation Datasets}

The process of collecting or creating evaluation datasets that adhere to the 5 D's can be broken down into three high-level approaches: benchmark analysis, \textit{golden} human annotated datasets, and synthetically-generated \textit{silver} datasets.

\subsubsection{Benchmark Analysis}

Publicly available benchmarks (e.g., \cite{pustejovsky2023towards, hendrycks2020measuring, guo2023owl, alam2024ctibench}) offer quick initial insights but often lack use-case specificity, may be contaminated with training data, and require careful auditing for quality and licensing.

\subsubsection{Human-Annotated Golden Datasets}

Human annotation is preferable for tasks requiring specialized knowledge not found in foundation models, as it can reveal subtleties in what constitutes a high-quality response (e.g., correctness, usefulness, completeness). However, this process is often difficult, costly, and time-consuming, risking low-quality or biased annotations due to the need for deep domain expertise and team coordination. Common collection workflows include leveraging in-house Subject Matter Experts (SMEs), outsourcing to third parties, employing user data and feedback, or gathering prompts through User Experience Research (UXR) surveys. These methods require clear instructions to ensure data quality, and practitioners should be aware that user data may have privacy constraints and that feedback tends to be biased toward negative experiences. To streamline these workflows and improve the quality of the final "golden" datasets, specialized data labeling and annotation platforms \cite{Labelbox, SageMakerGroundTruth, Prodigy, doccano}  can help enforce consistency and manage annotator agreement.

\subsubsection{Synthetically-Generated Silver Datasets}

Synthetic "silver" datasets offer a scalable, cost-effective alternative to human annotation that can bypass potential legal and privacy issues. This approach, however, requires careful human oversight to maintain data quality, minimize bias, and ensure the data is decontaminated from the model's original training set. Common generation techniques are as follows.

{\bf Distilling from Frontier Models:} This involves using powerful foundation models to generate prompt-response pairs. Advanced methods \cite{wang2022self, wei2022chain,taori2023alpaca,chiang2023vicuna,langchain,llama_index,es2023ragas} can be used to guide this process, such as having the model self-critique its responses based on a set of principles (Constitutional AI) \cite{bai2022constitutional} or iteratively making prompts more complex or diverse (Evol-Instruct) \cite{bai2022evol}.

{\bf Increasing Diversity:} To ensure variety in the generated data, strategies include prompting the LLM to adopt different personas or using sampling techniques like adjusting the model's temperature or employing nucleus sampling (top-p) \cite{holtzman2019curious}.

{\bf Leveraging Automatic Data Scraping Pipelines:} Leveraging automated data scraping pipelines is a common method for amassing large volumes of domain-specific text from the internet or internal repositories. This process can be implemented using a range of tools, from building custom crawlers with open-source Python frameworks like Scrapy \cite{Scrapy} and Beautiful Soup \cite{BeautifulSoup}, to using browser automation libraries like Selenium \cite{Selenium} for dynamic websites, or employing commercial platforms for managed scraping. However, it is difficult to guarantee that data curated in this fashion is fully decontaminated. As mentioned earlier, it is critical to rigorously filter these web-based datasets. Various filtering methods can be applied, ranging from manual inspection to more scalable approaches. For instance, libraries like Pandas \cite{mckinney2011pandas} can be used to remove texts with low token counts, while LLMs or other lighter-weight machine learning models \cite{wolf-etal-2020-transformers} can be trained to identify and discard low-quality or irrelevant content.

\subsubsection{Curating Evaluation Dataset Metadata}
\label{sec:curating_eval_set_metadata}

Beyond prompts and responses, metadata can be curated for each data point.

{\bf Tags:} Descriptive text categories that can be manually or automatically applied to prompt/response pairs to measure and monitor dataset diversity.

{\bf Grounding Information:} Supporting context (e.g., a relevant news article) can be provided to an evaluation, particularly an autorater, to help it accurately judge factuality on topics that may be too recent or outside its training data. This information should not be sourced from the system being evaluated.

{\bf Expected Information:} Specific terms, steps, or fields that the model is required to recall in its response. Unlike tags, this information requires auditing for correctness, as its fidelity can directly impact certain evaluation metrics.

\subsubsection{When to Use Which Data Curation Approach}

Benchmark, human-annotated, and synthetically-generated datasets should complement one another. Starting with benchmarks or human-annotated data points, evaluators can easily generate more synthetic data points from there to glean directional insights on the task at hand. This bootstrapped dataset can be iteratively improved over time to meet the \textit{5 D's}, effectively turning a \textit{silver} dataset into a \textit{golden} one.

\subsection{Adhering to and Quantifying the 5 D’s}

Beyond relying on performance metrics, dataset quality can be assessed by quantitatively measuring adherence to the 5 D's. Consistently high performance may signal a weak dataset, so it is crucial that the dataset evolves with the metrics. Core strategies for measuring and adhering to each principle include the following.

{\bf 1. Demonstrative of Production Usage:} The evaluation dataset can be aligned with real-world use by classifying prompts into representative categories, incorporating user-reported bugs and correlating offline metrics with in-product user satisfaction surveys. A mismatch between high metric scores and low user satisfaction indicates the dataset is not representative.

{\bf 2. Diverse}:  Diversity can be measured by categorizing data along relevant dimensions (e.g., topic, difficulty), grouping similar prompts (e.g., (e.g., using LSH clustering \cite{indyk1998approximate} or embeddings \cite{rudd2024efficient})), and using n-gram or embedding-based metrics to quantify the similarity between data points.

{\bf 3. Defined Scope:} Defined Scope can be achieved by creating tailored datasets for individual system components (e.g., grounding APIs, tool use) in addition to the end-to-end system. This modular approach, analogous to unit testing, helps isolate the source of failures.

{\bf 4. Decontaminated:} Dataset contamination is caused by overlapping training\footnote{LLM training can consist of multiple stages including foundation pre-training, continued pre-training, supervised fine-tuning, preference alignment, and others - by \textit{training datasets}, we refer to the data used within any and all of these stages.} and evaluation data, and can lead to inflated performance metrics and poor end-system design, model selection, and functionality. To prevent this, it is best practice to maintain a separate, uncontaminated evaluation dataset, ideally private and collected after the training cutoff. There are several ways to check for contamination, including continuation testing\cite{274574}, i.e., checking if the model can reproduce evaluation data from partial prompts or generate exact responses and examining response confidence by inspecting log-probabilities or perplexity (cf. Section \ref{sec:perplexity}). Assuming access to the training data (not always available), direct content comparisons between the evaluation set and the training set can be performed through exact string matching, substring matching, or hash comparisons. Efficient approaches can be employed for large data \cite{manber1993suffix, bloom1970space}.

{\bf 5. Dynamic:} The dataset should be kept current. Implementing a process to regularly audit, update, version control, and add new data points while removing irrelevant or outdated ones, can help ensure that the evaluation remains relevant over time.

\subsection{Required Evaluation Set Sample Sizes}

Datasets must be large enough to measure representative metrics, while not imposing significant computational or quota costs given the number of evaluations needed. We can derive a rough estimate of minimal evaluation set size $n$ required to estimate the performance of the model and adequately represent outcomes when deployed via standard statistical sample size calculation \cite{singh2014sampling}.
 
\begin{equation}
 n = z^2 \times \hat{m}(1-\hat{m})/\epsilon^2.
\end{equation} 

Here, $z$ is the $z$-score for a chosen confidence level, $\hat{m}$ is the expected metric score (e.g., 0.8 for 80\% accuracy), and $\epsilon$ is the desired margin of error. For example, to achieve 95\% confidence ($z$=1.96) with a 5\% margin of error for a metric expected to be 80\%, approximately 246 samples would be required. Notably, the required sample size increases dramatically as the desired margin of error decreases.
 
\subsection{Wrap-Up: Connecting Datasets to the Evaluation Framework}

The formulation of evaluation datasets serves as the first foundational pillar in the proposed evaluation framework. The approaches detailed in this section -— from using public benchmarks to creating human-annotated or synthetically-generated datasets -— are all tools to achieve the core principle of a dataset that is demonstrative, diverse, decontaminated, dynamic, and defined in scope. A well-curated dataset, tailored to the specific goals that go into the system (Defined Scope and Objectives in Figure \ref{fig:teaser}), is a prerequisite for the subsequent selection of meaningful metrics, forming the empirical bedrock upon which the entire evaluation rests.

%% file: content/metrics.tex
\section{Metrics Selection}

Evaluation metrics, like datasets, should align with the evaluation's scope and objectives. 
Moreover, datasets must contain proper data to support metric calculation, e.g., some metrics require ground truth responses while others do not.
Different metrics are better suited for assessing specific qualities, such as factuality, fluency, and summarization/translation quality.
Since there is no single "silver bullet" metric, multiple metrics should be tracked to form a holistic understanding of system performance. 
The choice of metrics also depends on the dataset, as some require ground-truth references while others do not.

\begin{table*}[h!]
\centering
\caption{Example Prompt, Golden Response, and Set of Candidates}
\label{tab:llm-examples}
\renewcommand{\arraystretch}{1.5} 
\begin{tabularx}{\textwidth}{| >{\bfseries}l | X |}
\hline
\rowcolor{headercolor}
Component & Text \\
\hline
\cellcolor{promptcolor} Prompt & Please provide a brief summary of the primary purpose and key outcomes of the Apollo 11 mission. \\
\hline
\cellcolor{goldencolor} Golden Reference & The Apollo 11 mission's primary purpose was to achieve the national goal of landing humans on the Moon and returning them safely to Earth. Key outcomes included astronauts Neil Armstrong and Buzz Aldrin collecting the first lunar samples, deploying scientific experiments on the lunar surface, and demonstrating U.S. technological supremacy during the Cold War. \\
\hline
\cellcolor{errorcolor} Candidate A (Factually Inaccurate) & The Apollo 11 mission's primary purpose was to land humans on \textbf{Mars}. Key outcomes included astronauts Neil Armstrong and Buzz Aldrin collecting the first \textbf{Martian} samples and deploying scientific experiments on the \textbf{Martian} surface. \\
\hline
\cellcolor{paraphrasecolor} Candidate B (Correct Paraphrase) & The main goal of Apollo 11 was to successfully land a crew on the Moon and ensure their safe return. During the mission, the astronauts gathered moon rocks and set up science equipment, which was a major victory for the U.S. in the Cold War. \\
\hline
\cellcolor{incompletecolor} Candidate C (Incomplete) & The Apollo 11 mission landed Neil Armstrong and Buzz Aldrin on the Moon. \\
\hline
\cellcolor{irrelevantcolor} Candidate D (Irrelevant) & The Apollo 11 mission was a landmark achievement in aeronautics, leading to significant advancements in aircraft design and commercial airline safety protocols that are still in use today. \\
\hline
\end{tabularx}
\end{table*}

\subsection{Term Overlap Metrics}
\label{sec:term_overlap}

Term overlap metrics, have become de-facto standards throughout the literature. These compare generated text to golden reference samples. They are  simple, efficient, and language-agnostic, but have limitations. First, they often miss deeper semantic meaning and context and can yield substantially different scores when encountering paraphrases with identical semantics. Second, they are bad at addressing fluency, factuality, and structure. Finally they are sensitive to length and repetition. Thus, we recommend that metrics be aggregated over multiple diverse, well-written ground truth examples capturing the same semantics for each LLM response.

\subsubsection{ROUGE Metrics for Summarization}
\label{sec:rouge}

Recall-Oriented Understudy for Gisting Evaluation (ROUGE) \cite{lin2004rouge,lin2003automatic} is  a set of metrics which assesses the quality of automatically generated summaries by comparing overlapping units of text, like words, phrases, $n$-grams (ROUGE-$n$), longest common subsequence (ROUGE-L), and skip-bigrams (ROUGE-S) between a generated summary and a human-written reference summary. When computed based on recall, ROUGE scores emphasize capturing relevant information from the reference, but they might not penalize a summary for being too long. Precision based ROUGE, on the other hand, favors concise summaries, even if they miss some relevant details. ROUGE can also be computed based on $F_{\beta}$-Measure, weighting respective contributions between precision and recall according to $\beta$. ROUGE scores range from 0 to 1.\footnote{In practice, across a large dataset it is unlikely to have scores of 0 or 1 for either ROUGE or BLEU and often indicates an implementation error.} 

ROUGE scores aim to evaluate summarization quality, e.g., when an LLM is tasked with summarizing reports, logs, or briefs into shorter/different versions and there are golden samples of specific length and style with specific meaningful terms. ROUGE scores  are also useful when answering questions where the golden response contains multiple key terms and phrases that are considered mandatory for a "good" response. 

\noindent
{\bf Limitations:} ROUGE scores often capture only surface-level word/phrase similarity, but do not capture nuanced differences in meaning or context. This precludes accurate evaluation for factuality, and high scores can be achieved based on syntactic term overlap rather than underlying semantics, limiting utility in evaluating Q/A and  generation tasks that require semantic accuracy. Consider Candidate A from Table \ref{tab:llm-examples}. Because it shares most of its n-grams (like "Apollo 11 mission", "primary purpose", "astronauts Neil Armstrong and Buzz Aldrin") with the Golden Response, it would achieve a deceptively high ROUGE score despite not capturing factual accuracy. Conversely, ROUGE is sensitive to paraphrasing: Candidate B is factually correct, but would receive a lower ROUGE score than Candidate A because it uses synonyms ("main goal" vs. "primary purpose", "moon rocks" vs. "lunar samples"). This demonstrates ROUGE's limitations in handling valid paraphrasing, often rewarding syntax over substance, and highlights the need for multiple golden reference examples. In contexts where responses involve generation of code and scripts, ROUGE scores may also err on the side of syntax over substance.

\subsubsection{BLEU for Translation}

Bilingual Evaluation Understudy (BLEU) \cite{papineni2002bleu} is a set of metrics which measure translation quality, comparing a generated translation to a golden reference translation, measuring correspondence between $n$-grams, while penalizing brevity so that shorter translations are not over-rewarded. The brevity penalty is calculated based on the length of the candidate translation and the length of the reference translation. BLEU combines precision scores over several $n$-gram sizes with weights for each $n$-gram size. The larger $n$, the better ordering is taken into account within the BLEU score. Typically several values of $n$ are used. Like ROUGE, BLEU scores range from 0 to 1, with 1 indicating perfect match between translation and reference and 0 indicating no match.

BLEU scores were designed for evaluating translations, and something akin to BLEU may serve as a first order metric for generating code or domain-specific queries from natural language, but BLEU has limitations for these tasks: it can assess syntax and structure but cannot assess underlying functionality, logical correctness, or semantic meaning of generated code, nor does it deal well with comments/documentation. BLEU can also be used for evaluating generations that are short, where the emphasis is on precision. ROUGE should be chosen over BLEU for summarizing sequences, as BLEU may penalize valid rephrasing; ROUGE can offer insight into structure/content and in a recall regime does not penalize valid rephrasing.

Many of the aforementioned limitations of ROUGE extend to BLEU, as both of these are term-overlap metrics. However, with all term-overlap metrics, there is an inherent tradeoff between precision-based metrics and recall-based metrics. Consider Candidate C from Table \ref{tab:llm-examples}. A recall-oriented metric like ROUGE-L might score this reasonably well for capturing the main event, but a precision-focused metric like BLEU would penalize it for its brevity.

There are several spinoffs of BLEU. SacreBLEU \cite{post2018call} aims to standardize tokenization for BLEU. NIST \cite{doddington2002automatic} modifies the brevity penalty by reducing the influence of low frequency $n$-grams on the final score, and weights $n$-grams according to their information value so that the evaluation is not skewed by over emphasis of common $n$-grams. METEOR \cite{banerjee2005meteor} uses WordNet to accommodate variations in word forms and synonyms, utilizes both precision and recall, and introduces a fragmentation penalty to preserve word order. 

\subsubsection{Selection of Relevant Keywords}
One way to address the aforementioned "syntax over substance" errors (cf. Sec. \ref{sec:rouge}) is to compute term-overlap metrics between the candidate summary and selected terms, rather than computing them over entire candidate responses and golden reference responses. 
While keyword selection may improve evaluation it can also be brittle; e.g., repeating a question may result in good scores, even if the substance of the question is not addressed. 
It is therefore important to be thoughtful about when to use keyword selection, one’s choice of keywords, and what other metrics to combine it with.

\subsection{Semantic Similarity Metrics}
\label{sec:embeddings}

Semantic similarity metrics compare generated and reference responses by converting them into numerical vectors (embeddings) and calculating their similarity (e.g., using cosine similarity). Unlike heuristic approaches like TF-IDF \cite{salton1975vector} and BM-25\cite{robertson1995okapi}, embeddings are generated via machine learnt language models and are designed to capture syntax, semantics, and structure. Embedding models can range from lighter-weight, e.g., Word2Vec \cite{mikolov2013efficient} to heavier weight distillations of LLMs, e.g., \cite{lee2024gecko}.

Embeddings are straightforward to compute and provide quantifiable measures of semantic alignment, assuming that the embedding model captures underlying semantics of the data. A "good" embedding-based semnantic similarity metric would correctly identify that Candidate B from Table \ref{tab:llm-examples} is much closer in meaning to the Golden Response than Candidate A is. The vector embeddings resulting from  "Moon" and "lunar" would be close, while the embeddings resulting from  "Mars" and "Martian" would be distant, overcoming the simple term-overlap problem. For proprietary data formats, it may be necessary to train a custom embedding model. Moreover, embedding-based similarity does not directly address fluency or factual accuracy of the candidate response.

\begin{table*}[!h]
\centering
\setlength{\tabcolsep}{0pt} 
\setlength{\extrarowheight}{0pt}
\begin{tabular*}{\textwidth}{| >{\RaggedRight\arraybackslash}p{0.33\textwidth} | >{\RaggedRight\arraybackslash}p{0.33\textwidth} | >{\RaggedRight\arraybackslash}p{0.33\textwidth} |} 
\hline
\multicolumn{1}{|c|}{\textbf{Scenario 1}} & \multicolumn{1}{c|}{\textbf{Scenario 2}} & \multicolumn{1}{c|}{\textbf{Scenario 3}} \\ 
\hline
{\bf Description:} LLM takes a reference sample and is prompted whether responses A and B satisfy some criteria. The responses are binary (e.g., yes/no). & {\bf Description:} Rater takes a reference sample and is prompted whether responses A and B satisfy a certain criteria on a continuous scale (e.g., 0.0 to 1.0). & {\bf Description:} LLM takes a reference sample and is separately prompted whether responses A and B satisfy a certain criteria on a Likert scale (e.g., 1 to 5). \\
\hline
{\bf Test:} McNemar's Hypothesis test. Ratings for A and B, while independent of one another, are dependent on the reference sample. & {\bf Test:} Two-Tailed Paired T-Test, since this test compares the means of two related groups assuming continuity. & {\bf Test:} Wilcoxon Signed Rank Test. Intervals between scores can be unequal so there is no distributional guarantee; a non-parametric test is needed.\\ 
\hline
\end{tabular*}
\caption{Evaluation Scenarios and corresponding Hypothesis Tests for Autoraters.}
\label{table:scenarios_tests}
\end{table*}

\subsection{NLI/Entailment Metrics}
\label{sec:entailment}

Natural Language Inference (NLI a.k.a. entailment) metrics use ML models to determine the relationship between two statements: a premise (A) and a hypothesis (B). They assess if B logically follows from A (entailment), if A and B are unrelated (neutral), or if B contradicts A (contradiction). This is valuable for evaluating factuality by checking if a statement aligns with known facts.

Considering the examples in Table \ref{tab:llm-examples}, the Golden Response would be the premise. Candidate A would result in a "contradiction" score, as landing on Mars contradicts landing on the moon. Candidate B would result in an "entailment" score, as its claims logically follow from the premise. Candidate D would result in a "neutral" score, as its claims about airline safety are unrelated to the premise. This demonstrates how NLI is a powerful tool for evaluating factuality.

To implement NLI, a language model is fine-tuned on an entailment dataset to generate an entailment score.  The model can also incorporate additional context to guide the evaluation towards specific aspects like factual consistency or fluency.

As with all model-derived metrics, the quality of an entailment metric depends on the quality of the model used to derive the metric – e.g., the model must appropriately capture the syntax and semantics of the premise and the hypothesis.

\subsection{LLM Autorater Metrics}
\label{sec:ab_testing}

LLMs can serve as scalable automated raters ("autoraters") to assess response quality on various aspects like fluency, factuality, or safety. They can assign individual scores (point-wise) or compare multiple responses directly (side-by-side, or SxS). Autoraters can can handle nuance beyond simple metrics, for example,  consider the examples in Table \ref{tab:llm-examples}: if prompted to evaluate "factual accuracy", a properly functioning autorater would rank the responses: Golden > B > C > A > D. If prompted to evaluate only "fluency," it might rate all of them highly whereas if prompted to evaluate accuracy and completeness comparing B and C, the autorater would choose B. 

The simplest scenario is where an LLM takes a reference sample with responses A and B as context and is prompted to output a binary score 0/1 depending on whether it prefers response A or response B. More elaborate evaluation often involves analyzing SxS comparisons like an A/B test. Statistical hypothesis tests (cf. Table \ref{table:scenarios_tests}) can determine if one response is preferred over another with statistical significance (i.e., when the resulting $p$-value is below a chosen significance level, $\alpha$). However, a low $p$-value does not indicate the magnitude of the difference, so practical significance should also be considered and can be treated more formally by quantifying effect size.

Despite the utility of autoraters, an autorater's effectiveness depends on its own quality and any grounding context provided (cf. Sec. \ref{sec:curating_eval_set_metadata}). Key limitations include:

\begin{itemize}[noitemsep, topsep=0pt, partopsep=0pt, parsep=0pt]
    \item Resource intensiveness: LLM evaluation requires significant computation.
    \item Lack of standardization: Comparing results across different datasets and models is challenging.
    \item Biases: LLM raters can exhibit biases towards verbosity, position, self-enhancement, lists, and specific styles.
    \item Distorted aggregation: Simple win/lose ratings can obscure nuanced differences between responses.
\end{itemize}

\subsection{Perplexity}
\label{sec:perplexity}

Perplexity (PPL) \cite{jelinek1977perplexity} aims to estimate the probability of a response occurring, normalizing based on response length. PPL is a transformation on this such that lower PPL corresponds to higher probability. This simple calculation ostensibly measures the predictive power of a language model without requiring ground truth, but this is a very loose predictor of practical performance at best. Since it uses probabilities derived from the underlying model (not actual probabilities), it is susceptible to favoring overfit models. Considering Candidate D from Table \ref{tab:llm-examples}, the response is grammatically correct and uses plausible-sounding language. Therefore, the LLM that generated it might assign it a high probability (and thus a low perplexity score). This demonstrates clearly that perplexity is not a reliable measure of task performance or factual accuracy, as a model can be confident in a completely nonsensical answer.

\subsection{Wrap-Up: Connecting Metrics to the Evaluation Framework}

Metrics selection is the second foundational pillar in our evaluation framework, building upon the previously established objectives and datasets. As this section details, no single metric is a "silver bullet"; instead, a holistic understanding of system performance is formed by computing and tracking multiple different metrics. The key is to create a balanced scorecard depending on the application, complementing traditional term-overlap metrics like ROUGE  with semantic similarity, NLI and autorater-based evaluations to capture a more complete picture of quality. Once this suite of metrics is defined, it sets the stage for developing a robust methodology for executing the evaluation and generating these scores.

%% file: content/eval.tex
\section{Robust Evaluation Methodology \& Execution for Improving LLM-Reliant Systems}

In this section, we discuss how to incorporate specific methodologies into the design and implementation of evaluation suites to address the real-world challenges inherent in LLM-reliant systems. While prior sections focused on creating datasets and metrics, this section details the execution framework required to handle confounding factors like inherent non-determinism from stochastic processes, system sensitivity to minor input changes, and the difficulty of disentangling the effects of grounding data from the model's prior knowledge. Furthermore, we present strategies for measuring complex failure modes such as hallucinations and undesirable non-responses. By directly addressing these issues, practitioners can leverage evaluation outputs as actionable guidance throughout the development and deployment lifecycles of generative AI systems.
 
\subsection{Handling Non-Determinism and Sensitivity}

Sources of non-determinism in LLMs stem from stochastic computations, chained model calls, and grounding data/API inconsistencies. Understanding how the pattern of non-determinism occurs is important for making accurate, conservative estimates of system performance given the potential noise it can produce. Note that non-determinism may occur irregularly or over long time windows. For example, estimating system noise only in the evenings when traffic is low may miss the non-determinism that occurs when a single request is broken into multiple sub-tasks that are processed independently under higher load."

Self-Consistency \cite{wang2022self} is a repeated evaluation strategy for LLMs that can address non-determinism. It involves sampling multiple responses from similar prompts and selecting the most frequent. A common way to generate different responses is by setting the model's \textit{temperature} parameter to a value greater than zero to enable diverse sampling for the same prompt. Self-Consistency reduces non-determinism by providing the response the system is most likely to generate for a given prompt, and may improve response quality as well. However, it increases the number of LLM calls ($n$ $\cdot$ \# original calls), which may improve response quality but requires more resources and may adversely impact cost/latency.

Another way to address non-determinism is to estimate its effect directly, and adjust metrics estimates. For example, one could run the evaluation $n$-times on fixed inputs, average the results, and estimate the variability induced by the system noise. By employing a sufficiently large choice of $n$, e.g., 10 (or larger), one can then establish an ongoing baseline of system performance with reported variations, then compare performance impacting changes to this baseline to see if the metrics fall outside the error bounds produced by the inherent system noise.

LLM-based systems are sensitive to their inputs such that a small change in the wording, spelling, or even spacing of a prompt can lead to dramatic changes in the response. Thus, understanding how sensitive they are to small variations in the prompt is an important exercise, since these variations are likely to be presented by users in the wild. 

Sensitivity can be assessed by injecting various types of noise into the prompt and measuring how the responses change. This practice is often part of a broader strategy for behavioral testing \cite{ribeiro-etal-2020-beyond} or robustness evaluation \cite{wang2021textflint} and uses a range of perturbations to test model stability.
To make this actionable, a variety of noise injection techniques can be employed, ranging from semantically neutral changes to more destructive ones that alter the underlying meaning. For this type of sensitivity evaluation, example ways to introduce prompt noise include case-swapping, random whitespace insertion, random character swapping, and using an LLM to rewrite the prompts. Once these variations are created, one can evaluate the impact of this sensitivity on:

\begin{itemize}[noitemsep, topsep=0pt, partopsep=0pt, parsep=0pt]
\item Overall metrics: Identify metrics most affected by noise.
\item Eval set subgroups: Analyze sensitivity across different task types, topics, etc.
\item Individual data points: Identify data points with high noise susceptibility. These can be especially informative for discovering underlying issues with the prompts, or in the case of a larger composed LLM system, the upstream components (e.g., RAG difficulties, summarization failures, etc).
\end{itemize}

\subsection{Comprehensive Evaluation of LLM System Components}

Several frameworks (e.g., OneTwo \cite{onetwo2024github},  LangChain \cite{langchain}) exist for composing multiple LLM calls into sequences, with information contained in the output of one LLM call being used as input to another LLM call. Often, grounding data is retrieved and included as part of this process. Performance of these chained/composed systems is a function of the quality of their intermediate components (LLM calls, RAG/Gounding API calls, etc.). Thus, measuring quality of all intermediate outputs is important, as early-stage quality issues can induce downstream effects.

LLMs possess extensive prior knowledge from pre-training, biasing them to produce certain responses for a given input. In topic areas where they have substantial exposure, they may be able to produce high quality responses without grounding. However this reliance on prior knowledge can present difficulties when the LLM is faced with topic areas or questions it is less familiar with, leading to hallucinations. Ultimately, the purpose of grounding is to guide the model toward correct responses on topics that it knows less about. 

When grounding data is used, it is important to identify whether the correctness of the response comes from the provided context or a reliance on the model's prior knowledge. To do so, one must evaluate the relevance and completeness of the grounding data relative to the generated response. This can be operationalized through several methods: using human annotators with a scoring rubric, employing a separate autorater LLM to score the context's utility, or calculating automated metrics like the semantic similarity between the prompt and the grounding documents. If the grounding data is found to be incomplete or irrelevant through these checks, the LLM may be relying too heavily on its prior knowledge. Over-reliance can mask grounding issues, especially for specialized tasks where up-to-date grounding is essential. For further discussion of this tension, see \cite{gupta2024rag}.

Where feasible, one can evaluate the end-to-end system performance both with and without certain grounding components. For example, given a grounding extension \cite{lewis2020retrieval}, we could compute the same evaluation metrics both with and without grounding data – in the latter case providing no grounding context, then evaluate performance using end-to-end metrics. This approach is relatively simple and directly addresses whether adding or removing a component contributes to or diminishes the overall score. A disadvantage of ablation evaluations is that they do a bad job measuring individual component quality, when component interactions propagate downstream.   

To evaluate grounding modules directly, more traditional retrieval metrics can be used, including Precision ($P$), Recall ($R$), and $F_{\beta}$ at $k$, Mean Reciprocal Rank (MRR), Normalized Discounted Cumulative Gain (NDCG), and Mean Average $P/R/F_{\beta}$ over multiple retrieval depths.

\subsection{Hallucination \& Unhelpful Responses }

When factuality of LLM responses is critical (e.g., for Q/A tasks), directly estimating hallucination rates may be feasible by probing the LLM with fictitious entities. 
This can be accomplished by generating a series of questions with fictitious entities and observing if the LLM provides information about them or responds with "I don't know" or similar. These "I don't know" type responses can be identified heuristically or with a classifier for greater accuracy. For example, "Tell me about CVE-2037-1234567" should result in an "I don't know" or "does not exist" type of response (CVE-2037-1234567 does not exist because at the time of writing this 2037 is in the future).

Measuring an LLM's non-response rate is crucial, especially when answers exist and good grounding data is provided. An undesirable non-response occurs when there is sufficient grounding data and/or prior knowledge for the LLM to respond, but it fails to do so opting instead to reply with an "I don't know" type of response. This phenomenon can occur when LLM hallucination is an issue and prompt instructions are provided to reduce hallucination. There  is  an inherent  tension  between  controlling  hallucination  via  prompt  instructions, and  eliciting undesirable  non-response. Overly cautious prompts aimed at reducing hallucinations can inadvertently increase non-response rates. Instructions like "If you don't know the answer, simply say `I don’t know'" may cause a non-response even when sufficient grounding data is provided. 

To measure undesirable non-response in the Q/A setting, one can generate a dataset of prompts with questions about known (i.e. non-fictitious) entities and evaluate the frequency of "I don't know" or "does not exist" types of responses to them, either heuristically or ideally with an "I don't know" classifier. One can then analyze the rate at which the LLM fails to respond to questions that it should, given the grounding and expectations about prior knowledge, be able to answer and revise prompts as necessary.

\subsection{Wrap-Up: Connecting Methodology to the Evaluation Framework}

The strategies for robust evaluation and execution outlined in this section are the third foundational pillar in our evaluation framework, amd can be incorporated with the previously discussed datasets and metrics into a functioning evaluation suite. This section has discussed the practical methodologies required to handle the inherent challenges of LLM-reliant systems, such as non-determinism, prompt sensitivity, and the difficulty of measuring complex failures like hallucinations. By directly addressing these confounding factors, this pillar ensures the evaluation's outputs are reliable and can serve as the actionable guidance needed for Analysis and Iteration (see Figure \ref{fig:teaser}).

%% file: content/conclusion.tex
\section{Putting it All Together}

A successful evaluation strategy integrates the core pillars of this paper—Datasets, Metrics, and Methodology—into a single, iterative process. The process begins by defining clear objectives to guide the formulation of representative datasets. These datasets inform the selection of a balanced suite of metrics, and a robust methodology is then applied to execute the evaluation, addressing challenges like non-determinism, prompt sensitivity, and hallucinations. Ultimately, the results should be interpreted not as a final grade, but as actionable guidance for system improvements and decisions. This approach transforms evaluation from a one-off validation exercise into a continuous loop where the evaluation suite matures and evolves in lockstep with the AI system it measures.

%% file: content/acknowledgements.tex
\clearpage
\section*{Acknowledgments}

The authors would like to thank Cyrus Harvesf, Spencer Sugarman, and Scott Coull for their insightful comments and constructive feedback, which greatly improved this manuscript.